# Two-Stream Neural Networks for Tampered Face Detection

Peng Zhou*  Xintong Han*  Vlad I. Morariu  Larry S. Davis
University of Maryland, College Park
pengzhou@umd.edu {xintong,morariu,lsd}@umiacs.umd.edu

## Abstract

*We propose a two-stream network for face tampering detection. We train GoogLeNet to detect tampering artifacts in a face classification stream, and train a patch based triplet network to leverage features capturing local noise residuals and camera characteristics as a second stream. In addition, we use two different online face swapping applications to create a new dataset that consists of 2010 tampered images, each of which contains a tampered face. We evaluate the proposed two-stream network on our newly collected dataset. Experimental results demonstrate the effectiveness of our method.*

## 1. Introduction

People post photos every day on popular social websites such as Facebook, Instagram and Twitter. A considerable number of these photos are authentic as they are generated from people's real life and shared as a part of their social experience. However, maliciously or not, more and more tampered images, especially ones involving face regions, are emerging on the Internet. Image splicing, which is the most common tampering manipulation, is the process of cutting one part of a source image, such as the face regions, and inserting it in the target image. To make the tampered result more realistic, adjustments on the shape, boundary, illumination and scaling are necessary, which make tamper detection challenging. Given advances in face detection and recognition techniques, anyone is able to swap faces with low cost using mobile applications [2] or open-source software [1]. Some tampered image examples generated from commercial software are shown in Figure 1. Even after close inspection, people mistake the tampered faces as original, and the current face verification techniques [22, 24] will also determine the source face and the tampered face are from the same identity. The consequence would be even more serious if manipulated images are used for political or commercial purposes.

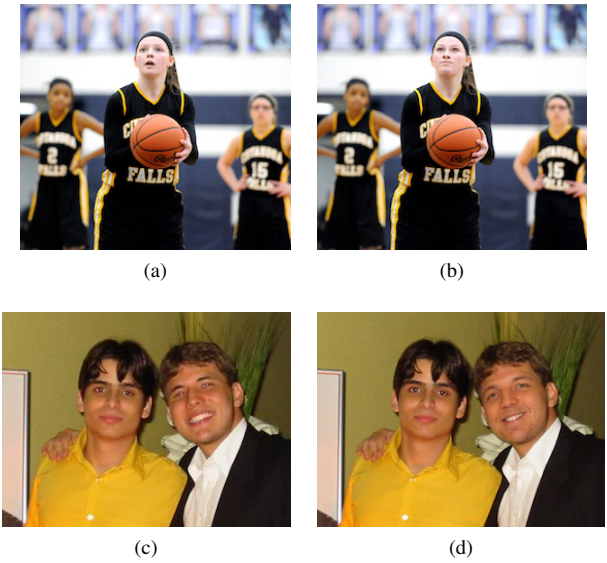

Figure 1. Examples of tampered faces. (a) original image. (b) Tampered image. The face in the middle has been tampered. (c) Original image. (d) Tampered image. The face on the right has been tampered.

Even though image tampering detection has been an active research area during the last decade, limitations still exist for current approaches since they focus on a particular source of tampering evidence [20, 14, 4, 11]. For example, local noise analysis fails to deal with tampered images constructed using careful post processing and Color Filter Array (CFA) models cannot deal with resized images.

To avoid focusing on specific tampering evidence and achieve robust tampering detection, we propose a two-stream network architecture to capture both tampering artifact evidence and local noise residual evidence as shown in Figure 2. This is inspired by recent research on CNNs showing the potential to learn tampering evidence [3, 21, 7] and rich models [15, 9, 10, 16], which are models of the noise components that produce informative features for steganalysis and we call the features produced by these mod-

*The first two authors contribute equally to this work.

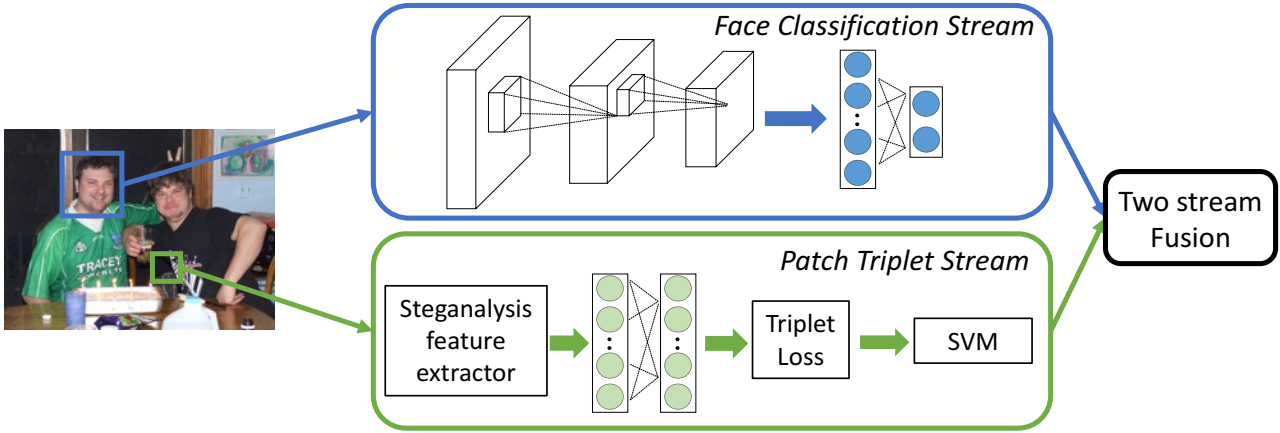

Figure 2. Illustration of our two-stream network. The face classification stream models visual appearance by classifying a face is tampered or not. The patch triplet stream is trained on steganalysis features to ensure patches from the same image are close in the embedding space, and an SVM trained on the learned features classifies each patch. Finally, the scores of two streams are fused to recognize a tampered face.

els "steganalysis features" in the rest of this paper, showing good performance on tampering detection. One of our streams is a CNN based face classification stream and the other one is a steganalysis feature based triplet stream. The face classification stream, based on GoogLeNet [23], is trained on tampered and authentic images and serves as a tampered face classifier. The patch triplet stream, based on patch level steganalysis features [15, 16], captures low-level camera characteristics like CFA pattern and local noise residuals. Instead of utilizing steganalysis features directly, we train a triplet network after extracting steganalysis features to allow the model to refine steganalysis features. Combining the two streams reveals both evidence of high-level tampering artifacts and low-level noise residual features, and yields very good performance for face tamper detection.

To train and evaluate our approach, we created a new face tampering dataset. The new dataset overcomes the drawback of existing datasets [19, 13, 12, 11] that are either small or in which the tampering is easily distinguishable even through visual inspection. We chose two face swapping apps to create two parallel sets of tampered images where the same target face is swapped with the same source face, but using different swapping algorithms. Only tampered images of good quality were retained. There are 1005 tampered images for each tampering technique (2010 tampered images in total) and 1400 authentic images for each subset. Evaluation on this dataset shows the effectiveness of our approach.

Our contribution is two-fold. First, by incorporating GoogLeNet and the triplet network as a two-stream network architecture, our method learns both tampering artifacts and local noise residual features. In addition, we create a new challenging dataset specific to face region tampering detection.

## 2. Related Work

There is growing research activity on tampering detection and localization. Prior methods can be classified according to the image features that they target, such as local noise estimation [20], double JPEG localization [4], CFA pattern analysis [14], illumination model [11] and steganalysis feature classification [8]. Recently, some methods based on Convolutional Neural Networks (CNN) [3, 7, 21] have achieved very good results.

The premise behind local noise estimation based techniques is that the difference between global noise characteristics and local noise characteristics reveals the hidden tampered regions. For example, Lyu *et al*. [18] cast noise statistic estimation as a closed-form optimization problem. By exploiting the property of kurtosis of natural images in band-pass domains and the relationship between noise characteristics and kurtosis, they reveal the inconsistency between global noise and local noise. However, the assumption that local noise is inconsistent with global noise fails if post processing techniques like filtering or image blending are applied after splicing. In contrast, our method is built on a triplet network to learn local noise residuals. This provides more reliable features for detection.

Double JPEG localization techniques can be classified into aligned double JPEG compression and non-aligned double JPEG compression, depending on whether or not the quantization factors align well after two applications of JPEG compression to the same image. The assumption is that the background regions undergo double JPEG compression while the tampered regions do not. For example, Bianchi *et al*. [4] presented a probability model that estimates DCT coefficients together with quantization factors. The advantage of this method is that it can be used for both aligned double JPEG and non-aligned double JPEG. However, this kind of technique strongly depends on the double JPEG assumption.

CFA localization based methods assume that the CFA pattern differs between tampered regions and authentic regions, due to the use of different imaging devices or other low-level artifacts introduced by the tampering process. By estimating the CFA pattern for the tampered image, it is possible to distinguish the tampered regions and authentic regions from each other. For example, Ferrara *et al*. [14] proposed an algorithm that estimates the camera filter pattern based on the fact that the variance of prediction error between CFA present regions (authentic regions) and CFA absent regions (tampered regions) is different. After a Gaussian Mixture Model (GMM) classification, the tampered regions can be localized. However, the assumption might not hold true if the tampered region has similar CFA pattern or the whole image is resized (whole resizing destroys the original camera CFA information and introduces new noise). Even though our method is also built on a CFA aware steganalysis features, we have a second stream that provides additional evidence in such scenarios.

Illumination based methods aim to detect illumination inconsistencies between tampered regions and authentic regions. For example, the tampered face and another face in the background may have different light source directions. Carvalho *et al*. [5] were able to estimate the light source direction for objects in an image, and thus use the light inconsistency to locate the tampered regions. De *et al*. [11] extracted the illumination features from image and use a Support Vector Machine (SVM) classifier for tampering classification. For face tamper detection, the performance can degrade as some applications only modify a small region of the tampered face, leaving the global illumination features of the face relatively unaltered.

Steganalysis feature [15] based methods extract diverse low-level information like local noise residuals. The steganalysis feature is a local descriptor based on cooccurrence statistics of nearby pixel noise residuals obtained from multiple linear and non-linear filters. Cozzolino *et al*. [8] used simplified steganalysis features and built a single Gaussian model to identify tampered regions. An improved method is [10], which treated tamper detection as anomaly detection and used a discriminative learning autoencoder outlier removing method based on steganalysis features. These methods show that steganalysis features are quite useful as low-level features which can be used in tamper detection. Goljan *et al*. [16] showed that this steganalysis model can also be utilized to estimate and extract CFA features, which extends their application. The difference between CFA aware steganalysis features and our approach is that we fine tune these features to improve their detection performance.

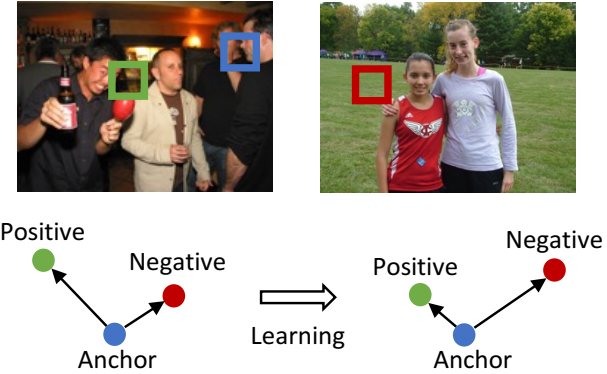

Figure 3. Illustration of weakly-supervised triplet network. By minimizing the triplet loss, the distance between patches from the same image (anchor and positive patches in the left image) in the learned embedding space becomes smaller than distance between two patches from different images (anchor patch in the left image and negative patch in the right image). Two boxes and circles of the same color represent a patch and its corresponding embedding, respectively.

Recently, methods based on CNNs have been developed. For example, by adding an additional median filter layer before the first convolutional layer, Chen *et al*. [7] achieved good performance in median filtering tampering. Bayar *et al*. [3] designed an adaptive filter kernel layer to estimate the filter kernel used in the tampering process, detecting various filtering tampered contents. However, the performance degrades significantly when multiple post processing techniques are applied to tampered regions. Rao *et al*. [21] combined a CNN with steganalysis features by initializing the kernel of the first convolution layer with steganalysis filter kernels.

## 3. Approach

Figure 2 shows our two stream framework. The face classification stream is a CNN trained to classify whether a face image is tampered or authentic. Thus, it learns the artifacts created by the tampering process. The patch triplet stream, which is trained on steganalysis features [16] of image patches with a triplet loss, models the traces left by in-camera processing and local noise characteristics.

### 3.1. Face Classification Stream

Since face tampering often creates artifacts (strong edges near lips, blurred areas on forehead, *etc*.), the visual information present in the tampered face plays an important role in tampered face detection. We adapt a deep convolutional neural network [23] trained for large-scale image recognition task, and fine-tune it to classify if a face is tampered or not. Given a face $q_i$, we denote the tampering score of this CNN as $F(q_i)$.

### 3.2. Patch Triplet Stream

In addition to modeling the visual appearance of tampered faces, we also leverage informative clues hidden in the in-camera processing for accurate tampered face detection. Recent research has shown that co-occurrence based local features (*e.g.*, steganalysis features [16]) can capture this hidden information and are effective in image splicing detection [10, 8]. In contrast to previous works that directly use these features, we refine the steganalysis features by a data-driven approach based on a triplet loss [22]. By training this triplet network, we ensure that a pair of patches from the same image are closer in the learned embedding space, while the distance between a pair of patches from two different images is large, as shown in Figure 3.

Formally, given an image patch $x_a$ (anchor patch), a patch $x_p$ (positive patch) from the same image, and $x_n$ (negative patch) from a different image, we enforce that the distance between $x_a$ and $x_p$ is smaller than that between $x_a$ and $x_n$ by some margin $m$:

$$d(f(r(x_a)), f(r(x_p))) + m < d(f(r(x_a)), f(r(x_n))) \quad (1)$$

where $r(x)$ is the steganalysis features of patch $x$, $f(r(x))$ is the embedding of $x$ we want to learn, and $d()$ is the sum of squares distance measure. We model $f$ by a two layer fully connected neural network.

This constraint is then converted into minimizing the following loss function:

$$L(f) = \sum_{a,p,n} \max(0, m + d(f_a, f_p) - d(f_a, f_n)) \quad (2)$$

where we use $f_a$ to denote $f(r(x_a))$ for simplicity. Instead of generating hard negatives in an online fashion [22], we randomly sample 15000 patch triplets from authentic images. Each triplet contains three $128 \times 128$ patches (one anchor, one positive, and one negative patch). We do not use hard negative sampling because our method is weakly supervised - for an anchor patch, its negative patches might be from the images taken from the same camera, thus have the same camera characteristics. During hard negative mining, these *pseudo* negatives will be treated as *true hard* negatives and the model will eventually project all patches into the same point in the embedding space in order to minimize $L$.

The triplet network is designed to determine whether or not two patches come from the same image. Face tampering detection works in a similar way. All the patches in an authentic region are from the same image and have small distances between each other in triplet embedding space, while the patches in tampered face regions have large distance from those authentic patches in the triplet embedding space because they are from another image. For each tampered image, the tampered regions have different characteristics from the authentic regions.

Therefore, we treat the tampered and authentic regions in each image as two different classes and train a classifier for each image to predict the tampered regions. We choose SVM as our classifier and train it for each test face on-the-fly. This process is shown in Figure 4. The SVM samples are obtained by extracting the triplet features on each patch through sliding windows. We treat the features extracted from non-face-region patches as negative samples because they are from the same image. To balance the negative samples, we randomly select the features extracted from other image patches as positive samples. Only the features from the automatically detected face regions in an image are treated as test samples. (*e.g.*, the left face of the test image in Figure 4). If a face region has been tampered, then the extracted features should have similar characteristics with positive samples; otherwise they should be similar to negative samples. For a patch $x$, the prediction of this SVM model $S(x)$ indicates the probability that $x$ is from another image, and thus is equivalent to the probability of tampering. As a face might contain multiple patches, we then take the average score for the patches in the face region as the final score for the face.

### 3.3. Two-stream Score Fusion

At test time, the final score for a face $q$ is obtained by simply combining the output scores of the two streams:

$$F(q) + \lambda \frac{1}{N_q} \sum_{x \in q} S(x) \quad (3)$$

where $N_q$ is the number of patches inside face $q$. $\lambda$ is a balance factor that ensures the two scores are at similar scale.

## 4. Experiments

In this section, we introduce our newly collected SwapMe and FaceSwap dataset, and then evaluate our method on it. Furthermore, we visualize the detection results of our method to better understand the proposed two-stream network. Finally, different training and test protocols are discussed.

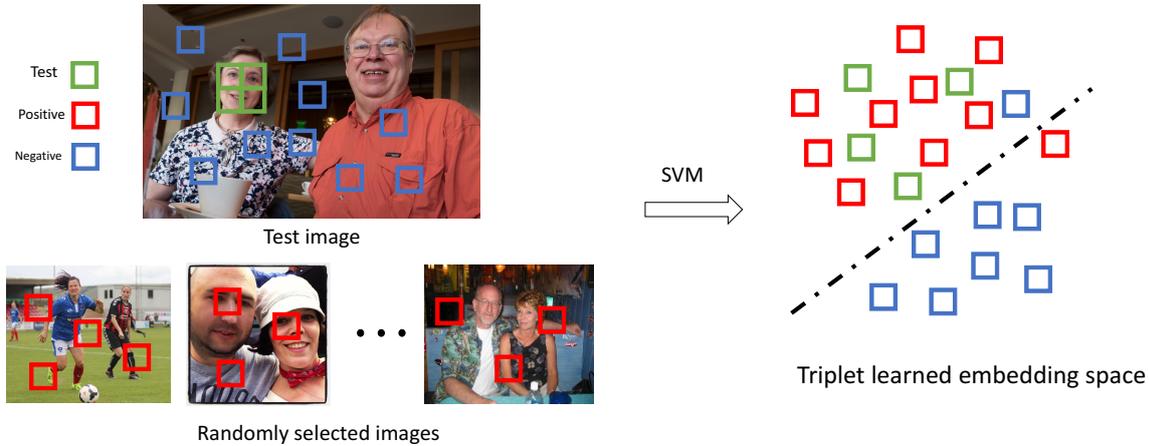

Figure 4. Demonstration of SVM training process. Suppose we want to test on the left face in the test image in a sliding window fashion. Green boxes are the test face patches; red boxes are randomly selected patches from other images and used as positive samples; blue boxes are the negative samples indicating patches from the same image. After SVM prediction, green boxes are more likely to be the ones from other images and thus the left face in the test image is classified to be a tampered face.

### 4.1. SwapMe and FaceSwap Dataset

Even though datasets for image tampering detection exist [19, 12, 13, 11], they are not well suited for large scale face tamper detection. Columbia Image Splicing dataset [19] and CASIA [12, 13] are large but most of the tampered regions are not human faces. DSI-1 dataset [11] focuses on face tampering but the total number of tampered images is only 25. Moreover, it is difficult to train deep learning methods on these datasets for face tamper detection.

To this end, we created a dataset utilizing one iOS app called SwapMe [2] and an open-source face swap application called FaceSwap [1]. Given a source face and a target face, they automatically replace the target face with the source face. Then, post processing such as boundary blurring, resizing and blending is applied to the tampered face, which makes it very difficult to visually distinguish the tampered from authentic images. Some examples created by SwapMe and FaceSwap are shown in Figure 5.

We selected 1005 target-source face pairs, and generated 1005 images with one tampered face in each image for each of the two applications. We further split these 1005 images into 705 for training and 300 for testing. The 705 training images together with another 1400 authentic images form the training set. The 300 test images are combined with 300 authentic images as the test set. For each test image, two faces are sampled for testing (one tampered and one authentic for a tampered image, and two authentic faces for an authentic image). Thus, in total, for each application, we have 705 tampered faces and 1400 authentic faces for training, and 900 authentic faces and 300 tampered faces for testing. Note that the selected images cover diverse events (*e.g.*, hol-idays, sports, conferences) and identities of different ages, genders and races. When needed, the face bounding boxes are generated using Dlib [17] face detection.

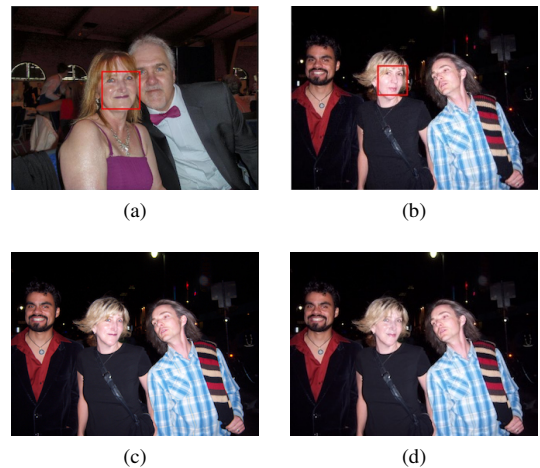

Figure 5. Examples of tampered images using SwapMe and FaceSwap. (a) Source image. The red bounding box shows the face moved to the tampered images. (b) Target image. The red bounding box shows the face before tampering. (c) Tampered image using SwapMe. (d) Tampered image using FaceSwap.

Our dataset has the following advantages: 1) It is a large dataset focus on face regions and is specifically designed for face tamper detection. 2) The quality of tampering is very good and the tampered faces look realistic. Generally, only face regions like the mouth, skin or eyes are tampered. 3) Since we use two different tampering techniques, we avoid learning the artifacts of one swapping algorithm, which may not be predictive when testing on the other.

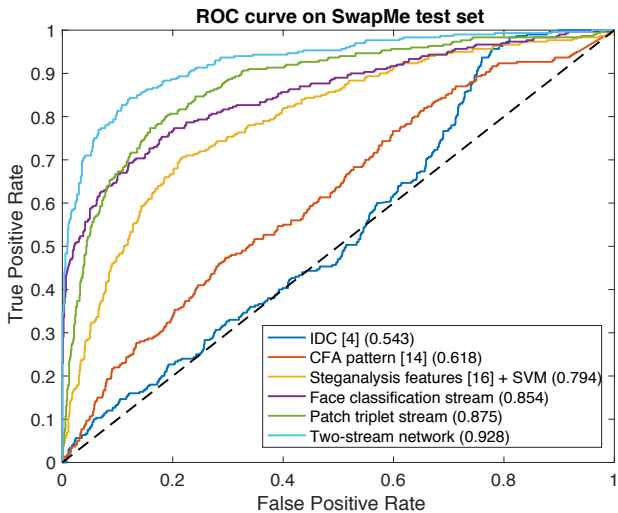

Figure 6. Face-level ROC comparison between our two-stream network and other methods.

| Methods | AUC |
|---|---|
| IDC [4] | 0.543 |
| CFA pattern [14] | 0.618 |
| Steganalysis features [16] + SVM | 0.794 |
| Face classification stream | 0.854 |
| Patch triplet stream | 0.875 |
| Two-stream network (Ours) | **0.927** |

Table 1. AUC of face-level ROC for different methods.

### 4.2. Experiment Setup

**Training and Test Protocol.** In order to avoid learning application-specific features, we train our model on one dataset and test on the other dataset. We train on FaceSwap training subset and test on SwapMe test subset. This is because tampering quality of FaceSwap is not as good as SwapMe. We do this for two reasons. On one hand, it might take attackers considerable effort to improve the tampered image in order to confuse the viewer. On the other hand, tamper detection algorithms typically will not know the specific technique attackers used to tamper images. We use the 705 FaceSwap tampered images and 1400 authentic images for training, and 300 SwapMe tampered images together with 300 authentic images for testing. The face-level tampering detection ROC curve and corresponding AUC are used for evaluation.

**Face Classification Stream.** We use GoogLeNet Inception V3 model [23] for training the tampered face classifier. Faces are resized to $299 \times 299$ and provided as input to the CNN. We set the initial learning rate to 0.1 and decrease it by a factor of 2 every 8 epochs. The batch size is set to 32. Finally we fine-tune all layers of the CNN pre-trained on ImageNet and stop the training process after 16k steps.

**Patch Triplet Stream.** Each triplet contains 3 $128 \times 128$ patches, and 5514D steganalysis features [16] are extracted for each patch. We randomly sample 15000 such triplets from authentic training images, of which 12000 are used for training and 3000 for validation. The triplet network contains two fully connected layers, the first layer contains 1024 neurons and the second one contains 512 neurons. The output of this network is then L2 normalized. During training, the initial learning rate is set to 0.1 and decreased by 2 every 8 epochs. The margin $m$ in Eqn. 1 is fixed to 0.04. During testing, we extract image patches in a sliding window fashion (window size = 128, stride = 64). The 512D learned representation is extracted for each patch then used for training a linear SVM using liblinear [6] with $C = 100$. Finally, we apply the trained SVM on face patches, and the average score of face patches is used as the face-level score.

### 4.3. Comparison with other methods

We compare our method with prior work on the SwapMe test set. The code for prior work is either provided by the authors or obtained from publicly available implementations on GitHub [1] [25]. The details of the methods and baselines are as follows.

**Face classification stream.** Only the output of the face classification CNN is used as the face tampering detection score.

**Patch triplet stream.** Only the patch triplet stream is used for detection of tampered faces.

**Steganalysis features [16] + SVM.** We train a linear SVM model directly on the features extracted from the steganalysis model [16]. This method is equivalent to removing the triplet network from our patch triplet stream.

**CFA pattern [14].** This method estimates the CFA pattern and uses a GMM algorithm to classify the variance of prediction error using the estimated CFA pattern. The output of this method is a local level tampering probability map. For the face region, an average probability is calculated as the final score.

**Improved DCT Coefficient (IDC) [4].** This method estimates the DCT coefficients for all the regions in the given image to find the singly JPEG compressed regions and classifies them as tampered regions. The output of this method is a probability map indicating tampering. To calculate the ROC curve, we take an average of the heat map score in the face region.

Figure 6 and Table 1 show the results on the SwapMe test set. CFA pattern [14] does not achieve good performance. Because images in the SwapMe dataset have been resized and the nearby pixel information is lost, this method fails to make correct predictions. The assumption of IDC [4] is that tampered regions are singly JPEG compressed while the authentic regions are doubly JPEG compressed. This is not the case in SwapMe, as both tampered regions and authentic regions have been doubly JPEG compressed. Steganalysis features + SVM performs reasonably well.

---
[1] https://github.com/MKLab-ITI/image-forensics

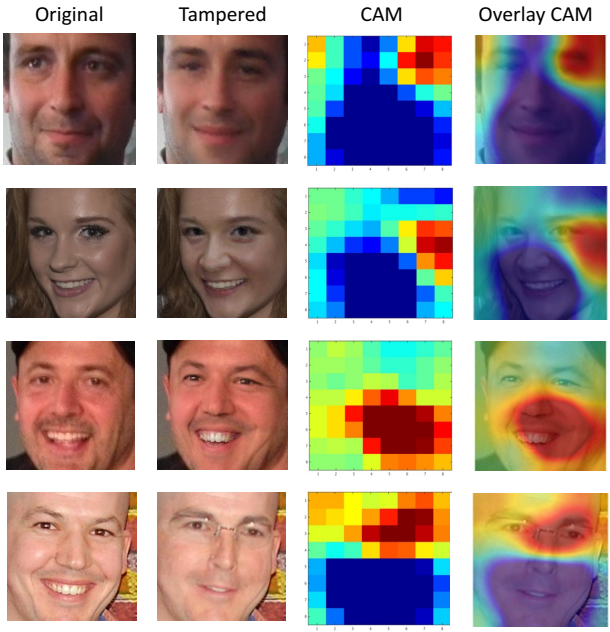

Figure 7. Class Activation Maps (CAMs) obtained from the face classification network. Each row shows the original image, the corresponding tampered face, the CAM, and a smoothing CAM overlaid with the tampered face for better visualization. In CAMs, red denotes high probability of tampering, and blue denotes low probability of tampering. We can observe that our face classification stream learns important artifacts created by the application during face tampering, such as stitching artifacts near face boundaries, strong edges around lips, and blurring effect when glasses are involved.

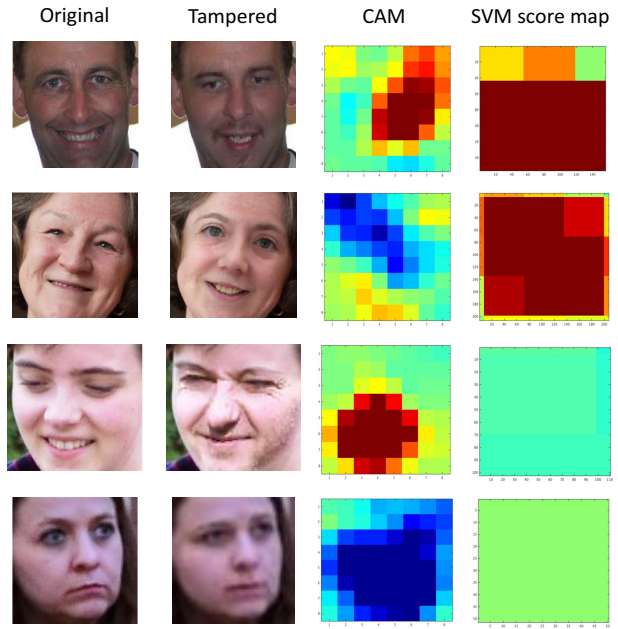

Figure 8. Heat map visualization of our two-stream network. Each row contains the original and tampered face, the corresponding CAM generated in the face classification stream in and SVM score map derived from the SVMs in the patch triplet stream. the In CAMs, red denotes high probability of tampering, and blue denotes low probability of tampering. In SVM score maps, red regions are more likely to be from different images other than the tampered images. In the first example, both streams can detect the tampered face. In the second and third examples, one stream fails while the other stream works, and fusing two streams successfully detects the tampered faces. Last row shows a failure case when the input face is small.

By refining steganalysis features, our patch triplet stream generates more informative features and obtains better result than steganalysis features + SVM (AUC improved from 0.794 to 0.875). By combining the patch triplet stream with the face classification stream, our full method models both high-level visual artifacts and low-level local features and thus is robust to post processing like resizing and boundary smoothing. As a result, it outperforms all other methods by a large margin.

### 4.4. Discussion

**Class Activation Map of Tampered Faces.** To better understand what visual clues the face classification stream relies on to detect a tampered face, we follow the method used in [26] to generate Class Activation Maps (CAMs) from the GoogLeNet, which are shown in Figure 7. Since the last feature map before the global average pooling layer in GoogLeNet is of size $8 \times 8 \times 2048$, the CAM for each face is of size $8 \times 8$. As shown in Figure 7, it is clear that our method learns the tampering artifacts created by the applications. The network is able to detect the stitching artifacts near the boundary of faces (as in the first two examples), strong edges near lips (as in the third example), and some blurring effect near eyes when glasses are involved during the tampering (last example). This visualization indicates that our approach is able to learn reasonable features that are useful for tampering detection.

**Effectiveness of Two-stream Fusion.** By fusing the detection scores of two streams, our method achieves better performance than each individual stream by a large margin. In Figure 8, we show some examples to visualize the detection results of both streams. In the first example, both streams can detect the tampered face - the face classifica-tion stream detects the unnatural edges near the mouth and beard, and the patch triplet network discovers the different noise residual distributions of the tampered region. In the second and third examples, only one stream is able to detect the tampering; however, with our two-stream fusion scheme, combining two streams detects the tampered face effectively. Our method fails to detect very small tampered faces as shown in the last example in Figure 8. In this example, the face is only of size $50 \times 50$. Since our face classification stream needs to resize the input face to $299 \times 299$, the upsampling of small faces loses some crucial visual information for tampering detection. Moreover, the patch size of our patch triplet stream is $128 \times 128$, which makes our method less robust when the tampered face is small because a large portion of the input patch will be authentic regions.

|                       | SwapMe test | FaceSwap test |
|-----------------------|-------------|---------------|
| SwapMe train          | 0.995       | 0.829         |
| FaceSwap train        | 0.854       | 0.998         |
| SwapMe + FaceSwap train | 0.995     | 0.999         |

Table 2. AUC of face classification stream comparison using different training and test splits. The row is the training dataset and the column is the test dataset.

**Tampered Face Detection in Different Protocols.** In addition to training on FaceSwap and testing on SwapMe, we report the results of different protocols in Table 2. The row is the training dataset and the column is the test dataset. We use 705 tampered + 1400 authentic images for training and 300 tampered + 300 authentic images for test when both training and testing are from the same dataset. We use 705 $\times$ 2 + 1400 images for training and 300 tampered + 300 authentic images for test when training on both SwapMe and FaceSwap and testing on one of them. The near perfect performance on either SwapMe or FaceSwap test set when training on common datasets indicates that our face classification stream has learned application-specific features.

## 5. Conclusion

We described a two-stream tampered face detection technique, where one stream detects low-level inconsistencies between image patches and another stream explicitly detects tamper faces. To evaluate our approach, we created a new data set that is challenging as multiple post processing are applied to the spliced region. The experimental results show that our approach outperforms other methods because it is able to learn both tampering artifacts and hidden noise residual features. Since our method needs steganalysis features extraction for the triplet network, an end-to-end architecture design is worth investigating and will be one of the main directions for future research.

## Acknowledgement

This work was supported by the DARPA MediFor program under cooperative agreement FA87501620191, "Physical and Semantic Integrity Measures for Media Forensics". The authors would like to thank Xi Yi for helping create the dataset and valuable discussions.